\documentclass[journal]{IEEEtran}
\usepackage[utf8]{inputenc}

\usepackage{amsmath, amsthm, amssymb, amsfonts}
\usepackage{algorithmic, algorithm}
\usepackage{mathrsfs}
\usepackage{comment}
\usepackage{enumitem}
\usepackage{float,xcolor}
\usepackage{hyperref}
\hypersetup{colorlinks,linkcolor={red},citecolor={olive},urlcolor={red}}

\newtheorem{theorem}{Theorem}

\ifCLASSINFOpdf
\else
\usepackage[dvips]{graphicx}
\fi
\usepackage{url}

\DeclareMathOperator*{\argmin}{arg\,min}
\DeclareMathOperator*{\tr}{tr}

\usepackage{graphicx}

\newcommand{\E}{\mathbb{E}}

\newtheorem{innercustomassumption}{}
\newenvironment{customassumption}[1]
{\renewcommand\theinnercustomassumption{#1}\innercustomassumption}
{\endinnercustomassumption}

\begin{document}
	
	\title{{\huge COVID-19 Time-series Prediction \\ by Joint Dictionary Learning and Online NMF}}
	\author{Hanbaek Lyu, Christopher Strohmeier, Georg Menz, and Deanna Needell
		%}
		%\author{First A. Author, \IEEEmembership{Fellow, IEEE}, Second B. Author, and Third C. Author, Jr., \IEEEmembership{Member, IEEE}
		\thanks{CS and DN supported in part by NSF CAREER \#1348721 and NSF BIGDATA \#1740325.}
		\thanks{Authors are with the Department of Mathematics, University of California, Los Angeles CA 90095 USA  (e-mail: \textcolor{red}{hlyu}@math.ucla.edu).}
		\thanks{Codes are available at \url{https://github.com/HanbaekLyu/ONMF-COVID19}}
		%\thanks{S. B. Author, Jr., was with Rice University, Houston, TX 77005 USA. He is now with the Department of Physics, Colorado State University, Fort Collins, CO 80523 USA (e-mail: author@lamar.colostate.edu).}
	}
	
	%\markboth{Journal of \LaTeX\ Class Files, Vol. 14, No. 8, August 2015}
	%{Shell \MakeLowercase{\textit{et al.}}: Bare Demo of IEEEtran.cls for IEEE Journals}
	\maketitle
	
	\begin{abstract}
		Predicting the spread and containment of COVID-19 is a challenge of utmost importance that the broader scientific community is currently facing. One of the main sources of difficulty is that a very limited amount of daily COVID-19 case data is available, and with few exceptions, the majority of countries are currently in the ``exponential spread stage," and thus there is scarce information available which would enable one to predict the phase transition between spread and containment. 
		
		In this paper, we propose a novel approach to predicting the spread of COVID-19 based on dictionary learning and online nonnegative matrix factorization (online NMF). The key idea is to learn dictionary patterns of short evolution instances of the new daily cases in multiple countries at the same time, so that their latent correlation structures are captured in the dictionary patterns. We first learn such dictionary patterns by minibatch learning from the entire time-series and then further adapt them to the time-series by online NMF. As we progressively adapt and improve the learned dictionary patterns to the more recent observations, we also use them to make one-step predictions by the partial fitting. Lastly, by recursively applying the one-step predictions, we can extrapolate our predictions into the near future. Our prediction results can be directly attributed to the learned dictionary patterns due to their interpretability.   
		
		%I think this can be omitted from the abstract
		%Our proposed method is very flexible and can be improved by augmenting additional COVID-19-related time-series. It also has the potential to reveal nontrivial correlations among other related time-series. These include spread of COVID-19 media information, medical and food supply shortages and demands, patient subgroup infections, and many more. 

	\end{abstract}
	

	\IEEEpeerreviewmaketitle

	\section{Introduction}
	The rapid spread of coronavirus disease (COVID-19) has had devastating effects globally. The virus first started to grow significantly in China and then in South Korea around January of 2020, and then had a major outbreak in European countries within the next month, and as of April the US alone has over 400,000 cases with over 12,000 deaths. Predicting the rapid spread of COVID-19 is a challenge of utmost importance that the broader scientific community is currently facing.  
	
	A conventional approach to this problem is to use \textit{compartmental models} (see, e.g.~\cite{keeling2005networks, brauer2008compartmental} ), which are mathematical models used to simulate the spread of infectious diseases governed by stochastic differential equations describing interactions between different compartments of the population (e.g.~susceptible, infectious, and recovered). Namely, one may postulate a compartmental model tailored to COVID-19 and find optimal parameters for the model by fitting it them the available data. An alternative approach is to use data-driven machine learning techniques, especially deep learning algorithms \cite{deng2013recent,bengio2013deep,deng2014tutorial}, which have had great success in various problems including image classification, computer vision, and voice recognition \cite{krizhevsky2012imagenet, boureau2010theoretical, hannun2014deep,amodei2016end}. 
	
	In this paper, we propose an entirely different approach to predicting the spread of COVID-19 based on \textit{dictionary learning} (or \textit{topic modeling}), which is a machine learning technique that is typically applied to text or image data in order to extract important features of a complex dataset so that one can represent said dataset in terms of a reduced number of extracted features, or dictionary atoms \cite{steyvers2007probabilistic, blei2010probabilistic}. Although dictionary learning has seen a wide array of applications in data science, to our best knowledge this work is the first to apply such an approach to time-series data and time-series prediction.
	
	Our proposed method has four components:
	
	\begin{description}[labelindent=0.2cm]
		\item[1.] (\textit{Minibatch learning}) Use online nonnegative matrix factorization (online NMF) to learn ``elemental" dictionary atoms which may be combined to approximate short time evolution patterns of correlated time-series data. 
		\vspace{0.1cm}
		
		\item[2.] (\textit{Online learning}) Further adapt the minibatch-learned dictionary atoms by traversing the time-series data using online NMF. 
		
		\vspace{0.1cm}
		\item[3.] (\textit{Partial fitting and one-step prediction})  Progressively improve our learned dictionary atoms by online learning while concurrently making one-step predictions by partial fitting. 
		
		\vspace{0.1cm}
		\item[4.] (\textit{Recursive extrapolation}) By recursively using the one-step predictions above, extrapolate into the future to predict future values of the time-series.
	\end{description}
	
	%The central concept in our method is \textit{dictionary learning}, which is a machine learning technique that extracts a reduced number of important patterns (dictionary atoms) in a complex data set. 
	
	Our method enables us to learn dictionary atoms from a diverse collection of correlated time-series data (e.g.~new daily cases of COVID-19, number of fatal and recovered cases, and degree of observance of social distancing measures). The learned dictionary atoms describe ``elemental" short-time evolution patterns from the correlated data which may be superimposed to recover and even predict the original time-series data. \textit{Online Nonnegative Matrix Factorization} is at the core of our learning algorithm, which continuously adapts and improves the learned dictionary atoms to newly arrived time-series data sets. 
	
	There are a number of advantages of our proposed approach that may complement some of the shortcomings of the more traditional model-based approach or large-data-based machine learning approach. First, Our method is completely model-free and has the universality of data types, as the dictionary atoms directly learned from the data serve as the `model' for prediction. Hence a similar method could be applied to predict not only the spread of the virus but also other related parameters. These include the spread of COVID-19 media information, medical and food supply shortages and demands, patient subgroup infections, immunity and many more. Second, our method does not lose interpretability as some of the deep-learning-based approaches do, which is particularly important in making predictions for health-related areas. Third, our method is computationally efficient and can be executed on a standard personal computer in a short time. This enables our method to be applied in real-time in online-setting to generate continuously improving prediction. Lastly, our method has a strong theoretical foundation based on the recent work \cite{lyu2019online}.
	
	In this article, we demonstrate our general online NMF-based time-series prediction method on COVID-19 time-series data by learning a small number of fundamental time evolution patterns in joint time-series among the six countries in three different cases (confirmed/death/recovered) concurrently. Our analysis shows that we can indeed extract interpretable dictionary atoms for short-time evolution of such correlated time-series and use them to get accurate short-time predictions with a small variation. This approach could further be extended by augmenting various other types of correlated time-series data set that may contain nontrivial information on the spread of COVID-19 (e.g.~time-series quantifying commodity, movement, and media data).

	This paper is organized as follows. In Section \ref{section:algorithms}, we give a brief overview of dictionary learning by nonnegative matrix factorization, and provide the full statement of our learning and prediction algorithms. In Subsection \ref{subsection:data_Set}, we give a description of the time-series data set of new COVID-19 cases and discuss a number of pre-processing methods for regularizing high fluctuations in the data set. Then we discuss our data analysis scheme and simulation setup in Subsection \ref{subsection:scheme}. In the following subsections \ref{subsection:minibatch}-\ref{subsection:prediction}, we present our main simulation results. Finally, we conclude and suggest further directions in Section \ref{section:conclusion}.

	\section{Time-series Prediction by Online NMF}
	\label{section:algorithms}
	
	\subsection{Dictionary learning by nonnegative matrix factorization}
	
\begin{figure}[H]
		\centering
		\includegraphics[width=1 \linewidth]{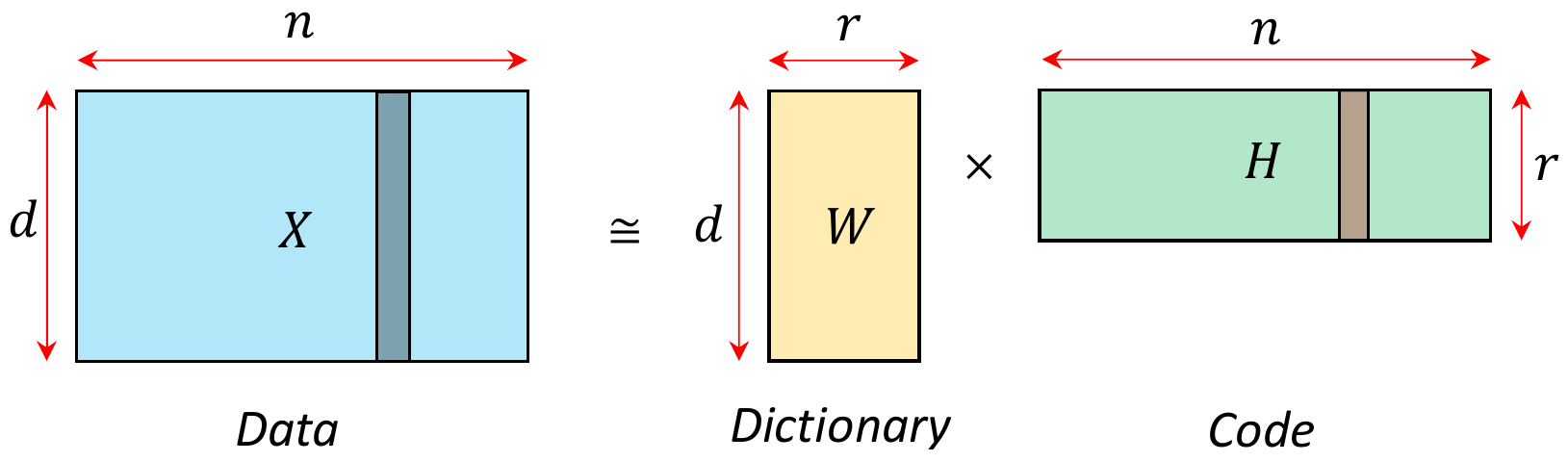}
		\caption{Illustration of  matrix factorization. Each column of the data matrix is approximated by the linear combination of the columns of the dictionary matrix with coefficients given by the corresponding column of the code matrix.}
		\label{fig:NMF_diagram}
	\end{figure}

	\textit{Matrix factorization} provides a powerful mathematical setting for dictionary learning problems. We first organize $n$ observations of $d$-dimensional samples a data matrix $X \in \mathbb{R}^{d \times n}$, and then seek a factorization of $X$ into the product $WH$ for some $W \in \mathbb{R}^{d \times r}$ and $H \in \mathbb{R}^{r \times n}$. This means that each column of the data matrix is approximated by the linear combination of the columns of the \textit{dictionary matrix} $W$ with coefficients given by the corresponding column of the \textit{code matrix} $H$ (see Figure \ref{fig:NMF_diagram}). This problem has been extensively studied under many names, each with different constraints: dictionary learning, factor analysis, topic modeling, component analysis. It has also found applications in text analysis, image reconstruction, medical imaging, bioinformatics, and many other scientific fields \cite{sitek2002correction, berry2005email, berry2007algorithms, chen2011phoenix, taslaman2012framework, boutchko2015clustering, ren2018non}.

	\textit{Nonnegative matrix factorization} (NMF) is an instance of matrix factorization where one seeks two nonnegative matrices whose product approximates a given nonnegative data matrix. Below we give an extension of NMF with an extra sparsity constraint on the code matrix which is particularly suited for dictionary learning problems \cite{Hoyer2004}. Given a data matrix $X\in \mathbb{R}^{d\times n}_{\ge 0}$, the goal is to find a nonnegative dictionary $W\in \mathbb{R}^{d\times r}$ and nonnegative code matrix $H\in \mathbb{R}^{r\times n}$ by solving the following optimization problem:
	\begin{align}\label{eq:NMF_error1}
	\inf_{W\in \mathbb{R}^{d\times r}_{\ge 0},\, H\in \mathbb{R}^{r\times n}_{\ge 0} }  \lVert X - WH  \rVert_{F}^{2} + \lambda\lVert H \rVert_{1},
	\end{align}
	where $\lVert A \rVert_{F}^{2} = \sum_{i,j} A_{ij}^{2}$ denotes the matrix Frobenius norm and $\lambda \ge 0$ is the $L_{1}$-regularization parameter for the code matrix $H$.

	A consequence of the nonnegativity constraints is that one must represent the data using the dictionary $W$ without exploiting cancellation. This is a critical mechanism that gives a parts-based representation of the data (see \cite{lee1999learning}). Many efficient iterative algorithms for NMF are based on block optimization schemes that have been proposed and studied following the introduction of the first, and most well-known, multiplicative update method by Lee and Seung \cite{lee2001algorithms} (see \cite{gillis2014and} for a survey).

	\subsection{Our algorithms}
	
	In this section, we provide algorithms for online dictionary learning and prediction for ensembles of correlated time-series. At the core of our online dictionary learning algorithm is the well-known online nonnegative matrix factorization (ONMF) algorithm \cite{mairal2010online, lyu2019online}, which is an online extension of NMF that learns a sequence of dictionary matrices from a sequence of data matrices. 
	
	We first illustrate the key idea in the simplest setting of time-series data for a single entity. Suppose we observe a single numerical value $x_{s}\in \mathbb{R}$ at each discrete time $s$. By adding a suitable constant to all observed values, we may assume without loss of generality that $x_{s}\ge 0$ for all $s\ge 0$. Fix integer parameters $k,N,r\ge 0$. Suppose we only store $N$ past data points at any given time, due to memory constraints. So at time $t$, we hold the vector  $\mathcal{D}_{t}=[x_{t-N+1},x_{t-N+2},\cdots,x_{t}]$ in our memory. The goal is to learn a dictionary of $k$-step evolution patterns from the observed history $(x_{s})_{0\le s \le t}$ up to time $t$. A possible approach is to form a $k$ by $N-k+1$ \textit{Hankel matrix} $X_{t}$ (see, e.g.,~\cite{rissanen1973algorithms}), whose $i$th column consists of the $k$ consecutive values of $\mathcal{D}_{t}$ starting from its $i$th coordinate. We can then factorize this into $k$ by $r$ dictionary matrix $W$ and $r$ by $t-k$ code matrices using an NMF algorithm:
	\begin{align}
	X_{t}=
	\begin{bmatrix}
	x_{t-N+1} & x_{t-N+2} & \cdots &  x_{t-k+1} \\
	x_{t-N+2} & x_{t-N+3} & \cdots & x_{t-k+2} \\
	\vdots & \vdots & \vdots &  \vdots \\
	x_{t-N+k} & x_{t-N+k+1} & \cdots & x_{t} \\
	\end{bmatrix}
	\approx
	W H.
	\end{align}
	This approximate factorization tells us that we can approximately represent any $k$-step evolution pattern from our past data $(x_{s})_{t-N< s \le t}$ by a nonnegative linear combination of the $r$ columns of $W$. Hence the columns of $W$ can be regarded as dictionary patterns for all $k$-step time evolution patterns in our current data set $\mathcal{D}_{t}$ for each time $t$.

	Below we provide an online mini-batch implementation of the above sketch of dictionary learning for time-series data in an online setting, as well as an online prediction algorithm.
	
	\begin{algorithmic}
		
		\begin{algorithm}
			\caption{Online dictionary learning for time-series data}\label{alg:ONMF_temporal}
			
			\STATE \textbf{Input:} Time-series $(\mathbf{x}_{t})_{1\le t \le T}$, $\mathbf{x}_{t}\in \mathbb{R}^{d\times 1}_{\ge 0}$
			\vspace{0.1cm}
			\STATE \textbf{Variables:}  $N\ge k\in \mathbb{N}$, $\lambda>0$, $\beta>0$, 
			\STATE \hspace{1.5cm} $\mathbf{W}_{0}\in \mathbb{R}^{d\times k \times r}_{\ge 0}$, $A_{0}\in \mathbb{R}^{r\times r}_{\ge 0}$, $B_{0}\in \mathbb{R}_{\ge 0}^{r\times dk}$
			
			\vspace{0.1cm}
			\FOR{$t = k,\cdots ,T$}
			\STATE \textit{Update sparse code}:
			\begin{align}\label{eq:sparse_coding_H}
			H_t = \argmin\limits_{H\in \mathbb{R}_{\ge 0}^{r\times (N-k+1)}} ||\mathbf{X}_t^{(3)} - \mathbf{W}_{t-1}^{(3)} H||_F^2 + \lambda||H||_1,
			\end{align}
			where $\mathbf{X}_{t}^{(3)}$ is the mode-3 unfolding of the $d\times k \times (N-k+1)$ tensor $\mathbf{X}_{t}$ (similarly for $\mathbf{W}_{t-1}^{(3)}$), whose mode-1 slices $\mathbf{X}_{t}(1),\cdots,\mathbf{X}_{t}(d)$ are given by the following Hankel matrix 
			\begin{align}
			\mathbf{X}_{t}(i)=
			\begin{bmatrix}
			\mathbf{x}_{t-N+1}(i) & \mathbf{x}_{t-N+2}(i) & \cdots &  \mathbf{x}_{t-k+1}(i) \\
			\mathbf{x}_{t-N+2}(i) & \mathbf{x}_{t-N+3}(i) & \cdots & \mathbf{x}_{t-k+2}(i) \\
			\vdots & \vdots & \ddots &  \vdots \\
			\mathbf{x}_{t-N+k}(i) & \mathbf{x}_{t-N+k+1}(i) & \cdots & \mathbf{x}_{t}(i) \\
			\end{bmatrix}.
			\end{align}
			\STATE \textit{Aggregate data}:
			\begin{align}
			A_t &= (1-t^{-\beta})A_{t - 1} + t^{-\beta} H_tH_t^T \label{e_A_t} \\ 
			B_t &= (1-t^{-\beta})B_{t - 1} + t^{-\beta} H_tX_t^T \label{e_B_t}
			\end{align}
			\vspace{-0.2cm}
			\STATE \textit{Update dictionary}:
			\begin{align}\label{e_W_t}
			\mathbf{W}_t = \argmin\limits_{\mathbf{W}\in \mathbb{R}_{\ge 0}^{d\times k\times r}} \tr(\mathbf{W}^{(3)} A_t (\mathbf{W}^{(3)})^T) - 2\,\tr(B_t \mathbf{W}^{(3)}) 
			\end{align}
			
			\ENDFOR
		\end{algorithm}
	\end{algorithmic}
	
	%%%%%%%%%%%%%%%%%%%%%%%%%%%%%%%%%
	
	\begin{algorithmic}
		
		\begin{algorithm}
			\caption{Partial fitting and Prediction}\label{alg:partial_fitting}
			
			\STATE \textbf{Input:} Time-series $(\mathbf{x}_{t})_{t-k+2\le s \le t}$, $\mathbf{x}_{t}\in \mathbb{R}^{d\times 1}_{\ge 0}$ 
			\vspace{0.1cm}
			\STATE \hspace{1cm} Dictionary tensor $\mathbf{W}_{t}\in \mathbb{R}^{d\times k \times r}$
			
			\STATE \textbf{Output:} Prediction $\hat{\mathbf{x}}_{t+1}$ for $\mathbf{x}_{t+1}$
			
			\vspace{0.1cm}
			\STATE \textbf{Variables:}  $\lambda'>0$
			
			\vspace{0.1cm}

			\STATE \textbf{Do:}
			
			\STATE \textit{Partial fitting:}
			\begin{align}\label{eq:partial_fitting_H_prediction}
			H_{*} = \argmin\limits_{H\in \mathbb{R}_{\ge 0}^{r\times 1}} ||\widetilde{\mathbf{x}}_{t} - \widetilde{W}_{t} H||_F^2 + \lambda'||H||_1,
			\end{align}
			where $\widetilde{\mathbf{x}}_{t}\in \mathbb{R}_{\ge 0}^{d(k-1)\times 1}$ is obtained by concatenating the columns of the following matrix 
			\begin{align}
			\begin{bmatrix}
			\mathbf{x}_{t-k+2}(1) & \cdots &\mathbf{x}_{t-k+2}(d) \\ 
			\vdots &\ddots &\vdots \\
			\mathbf{x}_{t-1}(1) & \cdots & \mathbf{x}_{t-1}(d) \\ 
			\mathbf{x}_{t}(1)  & \cdots & \mathbf{x}_{t}(d)
			\end{bmatrix},
			\end{align}
			and $\widetilde{W}_{t}\in \mathbb{R}_{\ge 0}^{d(k-1)\times r}$ is the mode-3 unfolding of the $d\times (k-1)\times r$ tensor $\mathbf{W}_{t}[:,:(k-1),:]$ that consists of the first $(k-1)$ entries of $\mathbf{W}$ in mode 2. 
			
			\vspace{0.1cm}
			\STATE \textit{Prediction:}
			\STATE $\hat{\mathbf{x}}_{t+1}=$ the last row of the $k\times d$ matrix obtained by reshaping $\mathbf{W}_{t}^{(3)} H^{*}$, where $\mathbf{W}_{t}^{(3)}$ is the mode-3 unfolding of $\mathbf{W}_{t}$
			
		\end{algorithm}
	\end{algorithmic}
	
	%%%%%%%%%%%%%%%%%%%%%%%%%%%%%%%%%%%%%%

	\begin{algorithmic}
		\begin{algorithm}
			\caption{Minibatch dictionary learning for time-series data}\label{alg:minibatch}
			
			\STATE \textbf{Input:} Time-series $(\mathbf{x}_{t})_{1\le t \le T}$, $\mathbf{x}_{t}\in \mathbb{R}^{d\times 1}_{\ge 0}$
			\vspace{0.1cm}
			
			\STATE \textbf{Output:} Dictionary tensor $\mathbf{W}_{M} \in \mathbb{R}^{d\times k \times r}_{\ge 0}$ and aggregate matrices $A_{M}\in \mathbb{R}^{r\times r}_{\ge 0}$, $B_{M}\in \mathbb{R}_{\ge 0}^{r\times dk}$
			\vspace{0.1cm}

			\STATE \textbf{Variables:}  $N\ge k\in \mathbb{N}$, $\lambda>0$, $\beta>0$, $M\in \mathbb{N}$
			
			\STATE \textit{Random initialization}: 
			\STATE \qquad Randomly initialize the tensors  $\mathbf{W}_{0}\in \mathbb{R}^{d\times k \times r}$, $A_{0}\in \mathbb{R}^{r\times r}_{\ge 0}$, $B_{0}\in \mathbb{R}_{\ge 0}^{r\times dk}$ so that the entries are i.i.d. $\textup{Uniform}([0,1])$ variables.
			
			\vspace{0.1cm}
			\FOR{$j=1,2,\cdots, M$}
			
			\STATE Choose $t$ uniformly at random from $\{T-N+1,T-N+2,\cdots,T\}$
			\vspace{0.1cm}
			
			\STATE Update $\mathbf{W}_{j}\leftarrow \mathbf{W}_{j-1}$ by using \eqref{eq:sparse_coding_H} - \eqref{e_W_t} in Algorithm \ref{alg:ONMF_temporal}.

			\ENDFOR
			
		\end{algorithm}
	\end{algorithmic}
	
	%%%%%%%%%%%%%%%%%%%
	
	The dictionary update rule in both Algorithms \ref{alg:ONMF_temporal} and \ref{alg:minibatch} are based on the well-known online NMF algorithm in \cite{mairal2010online, lyu2019online}. Furthermore, in Algorithms \ref{alg:ONMF_temporal} and \ref{alg:partial_fitting}, we also outline how to make predictions using the online-learned dictionary atoms via partial fitting (Algorithm \ref{alg:partial_fitting}) and extrapolation. Algorithm \ref{alg:minibatch} is useful for initializing the dictionary tensor $\mathbf{W}_{0}$ in Algorithm \ref{alg:ONMF_temporal}, especially when the time-series data is limited (small $T$) so that one may not expect the online dictionary learning in Algorithm \ref{alg:ONMF_temporal} in $T$ steps.  Also, we may use a different time-series $(\mathbf{y}_{t})_{1\le t \le T}$ to find a dictionary tensor $\mathbf{W}_{M}$ and use it as the initial dictionary for the given time-series $(\mathbf{x}_{t})_{1\le t \le T}$ in Algorithm \ref{alg:ONMF_temporal}. This ``transfer learning" would be effective when the two time-series share a similar structure.

	\begin{figure*}[ht]
		\centering
		\includegraphics[width=1 \textwidth]{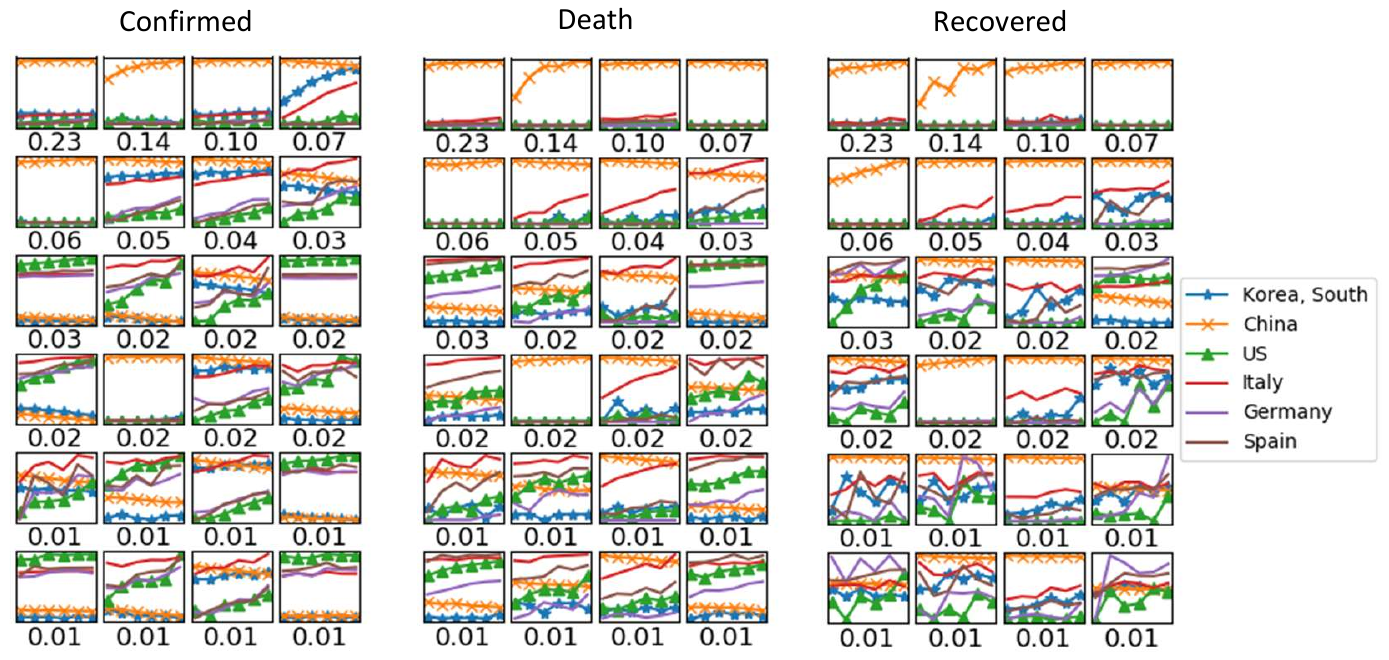}
		\caption{\small 24 Joint dictionary atoms of 6-day evolution patterns of new daily cases (confirmed/death/recovered) in six countries (S. Korea, China, US, Italy, Germany, and France). Each dictionary atom is a $6*6*3=108$ dimensional vector corresponding to $\texttt{time}*\texttt{country}*\texttt{case type}$. The corresponding importance metric is shown below each atom. 50 atoms are learned and the figure shows top 24 with the highest importance metric.}
		\label{fig:joint_dict}
	\end{figure*}
	
	The sparse coding problems in \eqref{eq:sparse_coding_H} and \eqref{eq:partial_fitting_H_prediction} can be solved by LASSO, whereas the constrained quadratic problem \eqref{e_W_t} can be solved by projected gradient descent algorithms. See \cite{lyu2019online} for more details and background. For practical use, we provide a python implementation of our algorithms in our GihHub repository (see the footnote of the front page).
	
	We also remark that, using the recent contribution of Lyu, Needell, and Balzano \cite{lyu2019online} on online matrix factorization algorithms on dependent data streams, we can give a theoretical guarantee of convergence of the sequence of learned dictionary atoms by Algorithm \ref{alg:ONMF_temporal} under suitable assumptions. The essential requirement is that the time-series data satisfies a weak ``stochastic periodicity" condition. A complete statement of this result and proof will be provided in our follow-up paper. 
	
	While this is a substantial extension of the usual independence assumption in the streaming data set, unfortunately, most COVID-19 time-series do not verify any type of weak periodicity condition directly, which is one of the biggest challenges in predicting the spread of COVID-19. In the next section, we describe our experiment setting and how we may overcome this issue of ``lack of periodicity" by combining the minibatch and online learning algorithms. It is important to note that minibatch learning algorithm (Algorithm \ref{alg:minibatch}) converges for fixed $T$ as $M\rightarrow \infty$, as it is a version of online NMF on a i.i.d. sequence of data, which satisfies our stochastic periodicity condition.

	\section{Application to COVID-19 time-series data sets}
	\label{section:application_covid_1}
	
	\subsection{Data set and pre-processing} 
	\label{subsection:data_Set}
	
	To illustrate our dictionary learning and prediction algorithms for time-series data, we analyze the historical daily reports of confirmed/death/recovered COVID-19 cases in six countries -- South Korea, China, US, Italy, Spain, and Germany -- from Jan 19, 2020 to Apr. 12, 2020. The input data can be represented as a tensor of shape $6\times 80 \times 3$ corresponding to countries, days, and types of cases, respectively\footnote{Raw data obtain from  \url{https://github.com/CSSEGISandData/COVID-19/tree/master/csse_covid_19_data}}. In order to apply our dictionary learning algorithms, we first unfold this data tensor into a matrix $X$ of shape $(d\times T) = (6\times 3) \times 80$, whose $t$th column, which we denote by $\mathbf{x}_{t}$, gives the full 18 dimensional observation in day $t, 0\le t\le 80$. Also, we find that the fluctuation in the original data set is too large to yield stable representations, let alone predictions. In order to remedy this, we pre-process the time-series data by taking a 5-day moving average and then entry-wise log-transform $x\mapsto \log(x+1)$. After applying dictionary learning and prediction algorithms, we take the inverse transform $x\mapsto \exp(x)-1$ and plot the result.
	%to plot the original and prediction.   

	\begin{figure*}
		\centering
		\includegraphics[width=1 \textwidth]{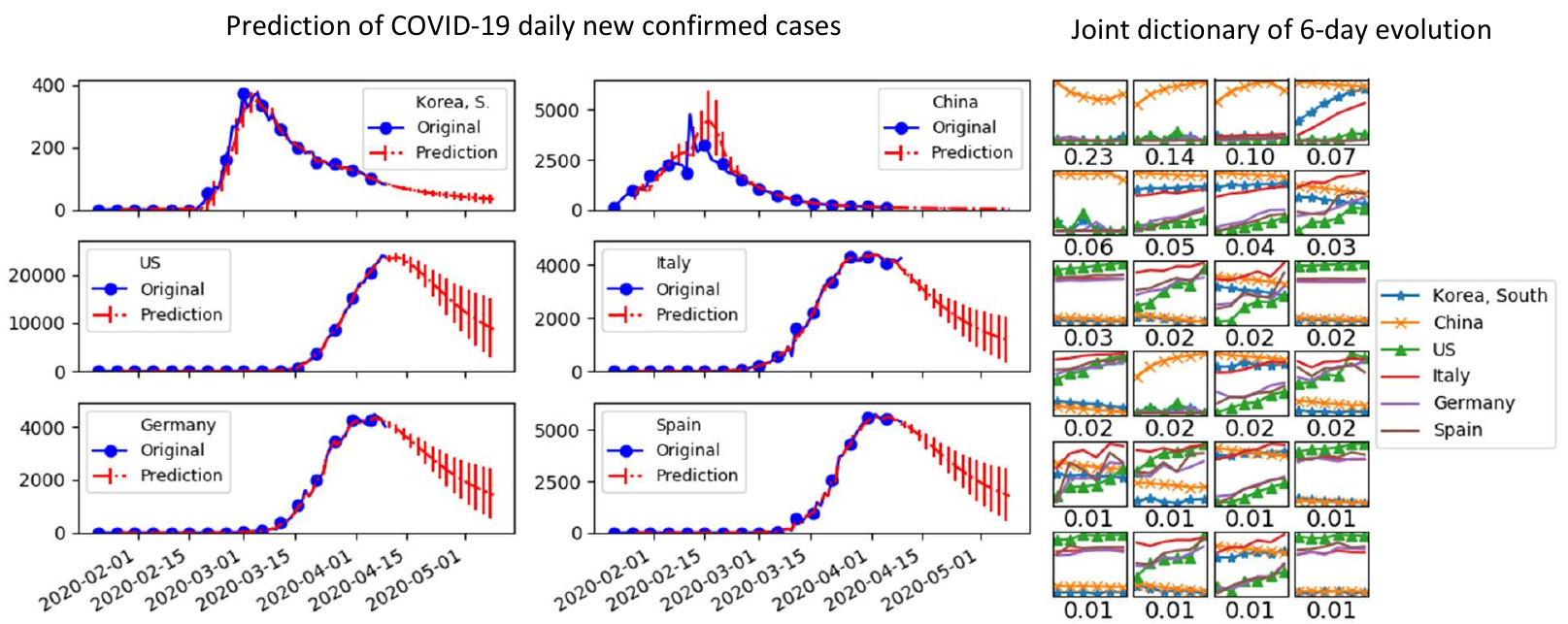}
		\caption{\small Joint dictionary learning and prediction for the time-series of new daily cases (confirmed/death/recovered) in six countries (S. Korea, China, US, Italy, Germany, and France). After joint dictionary atoms are learned by minibatch learning, they are further adapted to the time-series data by concurrent online learning and predictions. (Right) Joint dictionary atoms of 6-day evolution patterns of new confirmed cases. The corresponding importance metric is shown below each atom. (Left) Plot of the original and predicted daily new confirmed cases of the six countries. The errorbar in the red plot shows standard deviation of 1000 trials.}
		\label{fig:confirmed}
	\end{figure*}

	\subsection{Analysis scheme and experiment setup }
	\label{subsection:scheme}
	
	As the COVID-19 time-series data set we analyze here only consists of $T=80$ highly non-repetitive observations, applying the online dictionary learning algorithm (Algorithm \ref{alg:ONMF_temporal}) with random initialization is not sufficient for proper learning and accurate prediction. We overcome this by using the minibatch learning algorithm (Algorithm \ref{alg:minibatch}) for the initialization for the online learning and prediction. Namely, we use the following scheme: 
	\begin{description}
		\item[1.] (\textit{Minibatch learning}) Use minibatch Algorithm \ref{alg:minibatch} for the time-series $(\mathbf{x}_{t})_{0\le t \le T}$ to obtain dictionary tensor $\mathbf{W}_{M} \in \mathbb{R}^{d\times k \times r}_{\ge 0}$ and aggregate matrices $A_{M}\in \mathbb{R}^{r\times r}_{\ge 0}$, $B_{M}\in \mathbb{R}_{\ge 0}^{r\times dk}$. 
		
		\item[2.] (\textit{Online learning and one-step prediction}) Use the output in step 1 as the initialization for Algorithm \ref{alg:ONMF_temporal}. For each $t=1,\cdots,T$, iterates the steps in Algorithm \ref{alg:ONMF_temporal} as well as Algorithm \ref{alg:partial_fitting}. This outputs a dictionary tensor $\mathbf{W}_{T}$ and a prediction $(\hat{x}_{t})_{k\le t \le T+1}$. 
		
		\item[3.] (\textit{Recursive extrapolation}) For $T<t \le T+L$, recursively use Algorithm \ref{alg:partial_fitting} 
	\end{description}
	
	The hyperparameters we used in each steps above are given below:
	\begin{description}
		\item[1.] (\textit{Minibatch learning}) 
		\begin{align*}
		N &= 100 \qquad (\text{Memory size}) \\
		M &= 20 \qquad (\text{\# of minibatch iterations})\\
		k &= 6 \qquad (\text{segment length}) \\
		\lambda &= 3 \qquad (\text{$L_{1}$-regularizer of sparse coding in \eqref{eq:sparse_coding_H}})\\
		r &= 50 \qquad (\text{Number of dictionary atoms})\\
		\beta &= 1 \qquad \begin{pmatrix}\text{learning rate exponent for}  \\ \text{minibatch learning} \end{pmatrix}
		\end{align*}
		
		\item[2.] (\textit{Online learning and one-step prediction})
		\begin{align*}
		N &= 100 \qquad (\text{Memory size}) \\
		k &= 6 \qquad (\text{segment length}) \\
		\lambda' &= 1 \qquad (\text{$L_{1}$-regularizer of sparse coding in \eqref{eq:partial_fitting_H_prediction}})\\
		r &= 50 \qquad (\text{Number of dictionary atoms})\\
		\beta &= 4 \qquad \begin{pmatrix}\text{learning rate exponent for}  \\ \text{online learning} \end{pmatrix} \\
		\lambda' &= 0 \qquad (\text{$L_{1}$-regularizer of sparse coding in \eqref{eq:partial_fitting_H_prediction}})
		\end{align*}
		
		\item[3.] (\textit{Recursive extrapolation}) 
		\begin{align*}
		\lambda' &= 0 \qquad (\text{$L_{1}$-regularizer of sparse coding in \eqref{eq:partial_fitting_H_prediction}}) \\
		L &= 30 \qquad (\text{Future extrapolation length})
		\end{align*}
	\end{description}

	\begin{figure*}[ht]
		\centering
		\includegraphics[width=1 \textwidth]{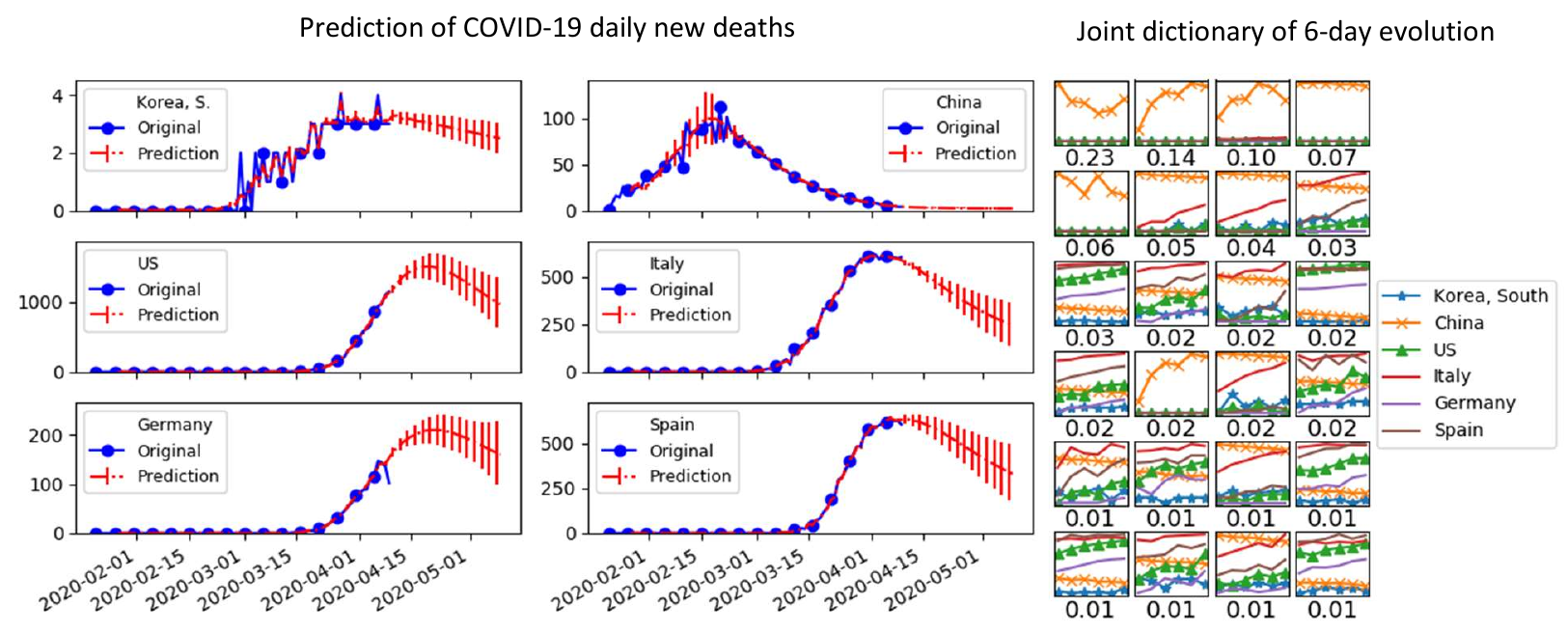}
		\caption{\small Joint dictionary learning and prediction for the time-series of new daily cases (confirmed/death/recovered) in six countries (S. Korea, China, US, Italy, Germany, and France). After dictionary atoms representing fundamental joint time-series patterns are obtained by minibatch learning, they are further adapted to the time-series data by online learning while making predictions. (Right) Joint dictionary atoms of 6-day evolution patterns of new death cases. The corresponding importance metric is shown below each atom. (Left) The plot of the original and predicted daily new death cases of the six countries. The error bar in the red plot shows the standard deviation of 1000 trials.}
		\label{fig:death}
	\end{figure*}

	\begin{figure*}[ht]
		\centering
		\includegraphics[width=1 \textwidth]{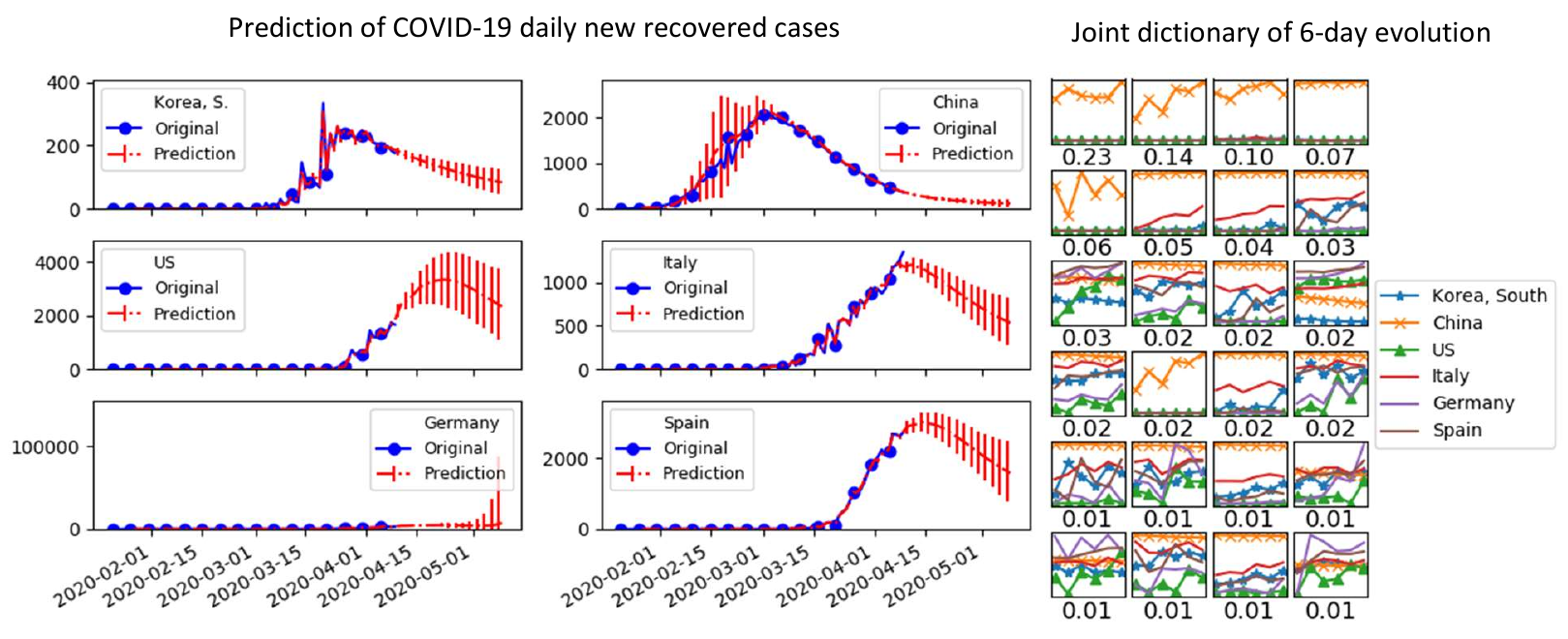}
		\caption{\small Joint dictionary learning and prediction for the time-series of new daily cases (confirmed/death/recovered) in six countries (S. Korea, China, US, Italy, Germany, and France). After joint dictionary atoms are learned by minibatch learning, they are further adapted to the time-series by online learning while making predictions. (Right) Joint dictionary atoms of 6-day evolution patterns of new recovered cases. The corresponding importance metric is shown below each atom. (Left) The plot of the original and predicted daily new recovered cases of the six countries. The errorbar in the red plot shows the standard deviation of 1000 trials. }
		\label{fig:recovered}
	\end{figure*}
	
	The role of the parameter~$\beta$ becomes clear when examining the equations~\eqref{e_A_t} and~\eqref{e_B_t}. It weights how much of the past data is used when updating the new dictionary matrix~$W_t$. For example, in  \eqref{e_W_t}, in the extreme case~$\beta=0$ the present observation at time $t$ is not used, wherease in the other extreme case~$\beta=\infty$ only the present observation is used.

	\subsection{Simulation results - Minibatch learning} 
	\label{subsection:minibatch}
	
	An example of dictionary atoms obtained from the minibatch learning algorithm (Algorithm~\ref{alg:minibatch}) from the COVID-19 daily new cases time-series data set is presented in Figure~\ref{fig:joint_dict}. We note that the time evolution structure in the data set is not used in this minibatch learning process, as slices of length $k$ evolution are sampled independently and uniformly at random from the entire history. Each dictionary atom is a $6*6*3=108$ dimensional vector corresponding to $\texttt{time}*\texttt{country}*\texttt{case type}$. 
	
	To each atom, we associate an ``importance metric" first introduced in \cite{lyu2019online} as a measure of the total contribution of the atom in representing the original data set. Namely, the importance metric of each atom is the total sum of its linear coefficients in the sparse coding problem \eqref{eq:sparse_coding_H} during the entire learning process. This is computed as the row sums of the sum of all code matrices $H_{t}$ in \eqref{eq:sparse_coding_H} obtained from the learning process.

	For example, the $(1,1)$ atom (in matrix coordinates) with importance 0.23 in Figure \ref{fig:joint_dict} indicates that the number of daily new confirmed cases in all six countries are almost constant and that China has significantly higher values than other countries. Also, the $(1,4)$ atom (in matrix coordinates) with importance 0.07 in Figure \ref{fig:joint_dict} indicates that the number of daily new confirmed cases are growing rapidly in Korea and Italy, while for the other four countries the values are almost constant. It is important to note that these dictionary atoms are learned by a nonnegative matrix factorization algorithm, so they maintain their individual interpretation we described before in representing the entire data set. 
	%which is also known as ``parts-based representation" of NMF \cite{lee1999learning}. 
	Indeed, such patterns were dominant in the COVID-19 time-series data during the period of Jan. 21 - Mar. 1, 2020 (see e.g.~the blue plot in Figure \ref{fig:confirmed}). Similarly, a direct interpretation of other dictionary atoms can be associated with features in the original data set.
	
	\subsection{Simulation results - Online learning and one-step prediction} 
	\label{subsection:online}
	
	The learned dictionary atoms in Figure \ref{fig:joint_dict} are not only directly interpretable but also provide a compressed representation of the original data set. Namely, we will be able to approximate any 6-day evolution pattern in the data set by only using the 24 atoms in Figure \ref{fig:joint_dict} with suitable nonnegative linear coefficients found from the sparse coding problem \eqref{eq:sparse_coding_H}. Obtaining a global approximation of the entire data set based on such local approximations by dictionary atoms is called \textit{reconstruction} (see for instance the image reconstruction example in \cite{lyu2019online}). Our partial fitting and prediction algorithm (Algorithm \ref{alg:partial_fitting}) extends this reconstruction procedure by using the learned patterns to make one step predictions adapted to the time-series data.

	The main point of our application is to illustrate that we may obtain reasonable predictions on our very limited $T=80$ COVID-19 time-series data set by learning a small number of fundamental patterns in joint time evolution among the six countries in three different cases (confirmed/death/recovered) concurrently. This approach could further be extended by augmenting various other types of correlated time-series which contain nontrivial information on the spread of COVID-19 (e.g.~time-series quantifying commodity, movement, and media data trends). One can think of learning the highly correlated temporal evolution patterns across different countries and cases as a model construction process for the joint time evolution of the data set. There are relatively few hyper-parameters to train in our algorithm compared to deep neural network-based models, and parameters in our method are in some sense built-in to the temporal dictionary atoms so that one does not need to begin by choosing the model to use.

	However, recall that the dictionary atoms learned by the minibatch algorithm are not adapted to the temporal structure of the data set. We find that further adapting them in the direction of time evolution by using our online learning algorithm (Algorithm \ref{alg:ONMF_temporal}) according to the scheme in subsection \ref{subsection:scheme} significantly improves the prediction accuracy by reducing the standard deviation of the prediction curves over a number of trials. An example of the result of this further adaptation of the minibatch-learned dictionary atoms is shown in Figures \ref{fig:confirmed}, \ref{fig:death}, and \ref{fig:recovered}. In each figure, the online-improved dictionary atoms are shown in the right, and the original time-series data and its prediction computed by Algorithms \ref{alg:ONMF_temporal}-\ref{alg:partial_fitting} are shown in blue and red, respectively. Prediction curves also show error bars of one standard deviation from 1000 trials of the entire scheme (minibatch + online + extrapolation) under the same hyperparameters in Subsection \ref{subsection:scheme}. 
	
	By comparing the corresponding dictionary atoms in Figures \ref{fig:joint_dict} and \ref{fig:confirmed} for example, we find that the online learning process does not change the importance metric on the top 24 atoms, and only a few atoms change in their shape significantly (especially the curves for China in atoms at $(1,1),(1,2),(1,3),(2,1),$ and $(4,2)$). Such new patterns were not able to be learned by the minibatch learning, but they were picked up by traversing the time-series data from the past to the present with our online learning algorithm. The ``correctness" of the learned dictionary atoms can be verified by the accuracy of the 1-step prediction up to time $T=80$ (the end of blue curve in Figure \ref{fig:confirmed}) in  Figures \ref{fig:confirmed}, \ref{fig:death}, and \ref{fig:recovered}.  

	We remark that our one-step predictions up to time $T=80$ are not exact predictions, as our initial dictionary tensor for this step, learned by the minibatch algorithm, uses all information in the entire time-series data. More precisely, one can think of this as a reconstruction procedure without seeing the last coordinate. This would have been a proper prediction if our initialization were independent of the data, but we find that online learning alone without the minibatch initialization gives inferior reconstruction results. We believe this is due to the fact that the COVID-19 time-series data set is too short (and far from periodic) for the online dictionary learning algorithm to converge in a single run. Nevertheless, in some sense the mini-batch method is implemented to compensate for the lack of data; given a sufficient amount of data, we could omit this step.

	\subsection{Simulation results - Recursive extrapolation} 
	\label{subsection:prediction}
	
	NextNext, we discuss the recursive extrapolation step, which gives the 30-day prediction of the new daily cases of all three types and all six countries simultaneously, shown as in Figures \ref{fig:confirmed}, \ref{fig:death}, and \ref{fig:recovered}. For instance, by partially fitting the first five coordinates of the length-6 dictionary atoms to the last five days of the data, we can use Algorithm \ref{alg:partial_fitting} to obtain a prediction of the future values at time $T+1$. We can then recursively continue this extrapolation step using the predicted data into the future ad infinitum. Our 30-day prediction results show reasonable variation among trials. However, the variation in prediction grows in the prediction length, so only a moderate range of predictions would have meaningful implications. 
	
	We remark that we did not enforce any additional assumption on the prediction curves (e.g.~finite carrying capacity, logistic-like growth for the total, SIR-type structure) that are standard in many epidemiological models. Instead, we chose our hyperparameters in subsection \ref{subsection:scheme} so that the future prediction curves approximately satisfy such assumptions, highlighting the model-free nature of our approach. This high-level fitting is not without compromise in the uncertainty of the prediction. For example, choosing large values of the $L_{1}$-regularization parameter $\lambda'$ used in the recursive extrapolation step reduces the variability of prediction significantly, but the prediction curve drops to zero very rapidly right after the end of the current data, which is absurd. 
	
	Lastly, we also mention that our recursive extrapolation defines a deterministic dynamical system in multidimensional space (dimension 18 in this case). The evolution is determined by the hyperparameters and the set of dictionary atoms at the end of the given time-series data ($T=80$ in this case). Even though our dictionary learning algorithms are randomzied, we find our 30-day predictions over many trials are within a modest standard error. However, we find that the mean trajectory of the prediction could vary significantly with respect to changes in the hyperparameters. Developing a more systematic method of choosing these hyperparameters is a future direction of inquiry.

	\section{Conclusion}
	\label{section:conclusion}
	
	With the rapidly changing situation involving COVID-19, it is critical to have accurate and effective methods for predicting short-term and long-term behavior of many parameters relating to the virus. In this paper, we proposed a novel approach that uses dictionary learning to predict time-series data. We then applied this approach to analyze and predict the new daily reports of COVID-19 cases in multiple countries. Usually, dictionary learning is used for text and image data; often with impressive results. To our best knowledge, our work is the first to implement dictionary learning to time-series data.
	
	There are a number of advantages of our approach that may complement some of the shortcomings of the more traditional model-based approach or the large-data-based machine learning approach. First, our method is completely model-free as the dictionary atoms directly learned from the data serve as the `model' for prediction. Our approach also works with universal data types. For example, it could be applied to predict not only the spread of the virus but also other related parameters. These include the spread of COVID-19 media information, medical and food supply shortages and demands, subgroup infections, immunity and many more. Second, the method works with small data sets. Most machine-learning methods need either model-specific input or large data sets. Because COVID19 only appeared recently, there are no large data sets yet available. Third, the method does not lose interpretability as some of the deep-learning-based approaches do. This is particularly important when making predictions for health-related areas. The learned dictionary atoms are not only interpretable but also identify hidden correlations between multiple entities. Fourth, our approach uses only a few hyperparameters that are model-free and independent of the data set. Therefore our approach avoids the issue of over-fitting. Furthermore, the method is computationally efficient and can be executed on a standard personal computer in a short time. Hence, it could be applied in real-time or online-settings to generate continuously improving predictions. Lastly, the method has a strong theoretical foundation: Convergence of the minibatch algorithm is always guaranteed and the convergence of the online learning algorithm is guaranteed under the assumption of quasi-periodicity. Therefore we expect our method to be robust i.e.~small changes to the data-set or to the parameters should not destroy the learning outcome.
	
	There are a number of future directions that we are envisioning. First, one could apply our method to county-level time-series data. One would obtain dictionary atoms describing county-wise correlation and prediction. This analysis could be critically used in distributing medical supplies and also in measuring the effect of re-opening the economy successively by a few counties at a time. Second, as not every individual can be tested, it is valuable to be able to transfer the knowledge on tested subjects to the ones yet to be tested. This `transfer learning' could naturally be done with our method, by learning a dictionary from one subject group and apply that to make predictions for the unknown group. Lastly, one may extend the method to a fully tensor-based setting, where a large number of related variables of different types could be encoded in a single tensor. Then one could use various direct tensor-factorization methods to learn higher-order dictionary atoms. For example, this might be useful in identifying a critical subgroup of variables for clinical data and therapeutics.

	\bibliographystyle{plain}
	\bibliography{bib.bib}
	
\end{document}